\documentclass{article}

\usepackage{microtype}
\usepackage{graphicx}
\usepackage{booktabs} % for professional tables
\usepackage{subcaption}
\usepackage{float}
\usepackage{hyperref}

\usepackage[accepted]{icml2024}

\usepackage{amsmath}
\usepackage{amssymb}
\usepackage{mathtools}
\usepackage{amsthm}

\usepackage[capitalize,noabbrev]{cleveref}

\theoremstyle{plain}

\theoremstyle{definition}

\theoremstyle{remark}

\usepackage[disable,textsize=tiny]{todonotes}
\newcommand{\papertitle}{Efficient Probabilistic Modeling of Crystallization at Mesoscopic Scale}

\icmltitlerunning{\papertitle}

\begin{document}

\twocolumn[
\icmltitle{\papertitle}



\icmlsetsymbol{equal}{*}

\begin{icmlauthorlist}
\icmlauthor{Pol Timmer}{tue}
\icmlauthor{Koen Minartz}{tue}
\icmlauthor{Vlado Menkovski}{tue}
\end{icmlauthorlist}

\icmlaffiliation{tue}{Department of Mathematics and Computer Science, Eindhoven University of Technology, Eindhoven, Netherlands}

\icmlcorrespondingauthor{Pol Timmer}{l.c.timmer@student.tue.nl}
\icmlcorrespondingauthor{Koen Minartz}{k.minartz@tue.nl}
\icmlcorrespondingauthor{Vlado Menkovski}{v.menkovski@tue.nl}

\icmlkeywords{Machine Learning, Neural Networks, Simulation, Probabilistic Models, Crystallization, Solidification, Mesoscopic Scale, VAE, PNS, Snow Crystals, Polycrystalline Solidification, Crystal Growth, Scaling}

\vskip 0.3in
]



\printAffiliations{}

\begin{abstract}
Crystallization processes at the mesoscopic scale, where faceted, dendritic growth, and multigrain formation can be observed, are of particular interest within materials science and metallurgy. These processes are highly nonlinear, stochastic, and sensitive to small perturbations of system parameters and initial conditions. Methods for the simulation of these processes have been developed using discrete numerical models, but these are computationally expensive. This work aims to scale crystal growth simulation with a machine learning emulator. Specifically, autoregressive latent variable models are well suited for modeling the joint distribution over system parameters and the crystallization trajectories. However, successfully training such models is challenging due to the stochasticity and sensitivity of the system. Existing approaches consequently fail to produce diverse and faithful crystallization trajectories. In this paper, we introduce the Crystal Growth Neural Emulator (CGNE), a probabilistic model for efficient crystal growth emulation at the mesoscopic scale that overcomes these challenges. We validate CGNE results using the morphological properties of the crystals produced by numerical simulation. CGNE delivers a factor of 11 improvement in inference time and performance gains compared with recent state-of-the-art probabilistic models for dynamical systems. 
\end{abstract}

\section{Introduction}
\begin{figure*}[h]
    \centering
    \begin{minipage}[b]{0.24\linewidth}
        \centering
        \includegraphics[width=\linewidth]{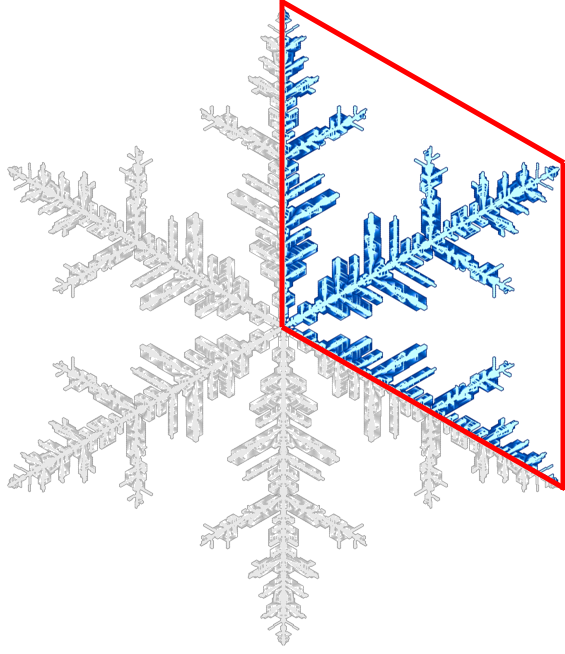}
        \subcaption{Original crystal representation on a hexagonal grid.}
        \label{fig:snowflake-symmetry-hex}
    \end{minipage}
    \hfill
    \begin{minipage}[b]{0.24\linewidth}
        \centering
        \includegraphics[width=\linewidth]{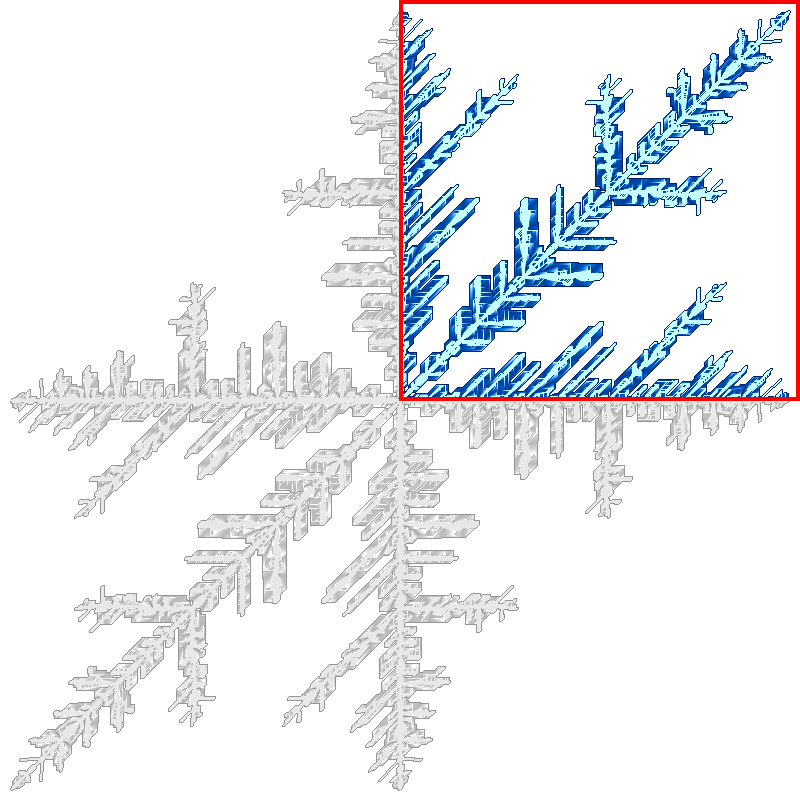}
        \subcaption{Transformed crystal representation on a square grid.}
        \label{fig:snowflake-symmetry-square}
    \end{minipage}
        \hfill
    \begin{minipage}[b]{0.24\linewidth}
        \centering
        \includegraphics[width=\linewidth]{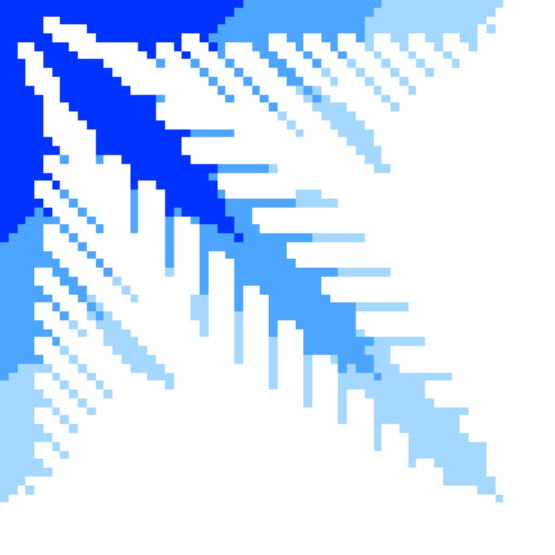}
        \subcaption{Crystal growth simulation by numerical LCA model.}
        \label{fig:snowflake-rollout-gt}
    \end{minipage}
    \hfill
    \begin{minipage}[b]{0.24\linewidth}
        \centering
        \includegraphics[width=\linewidth]{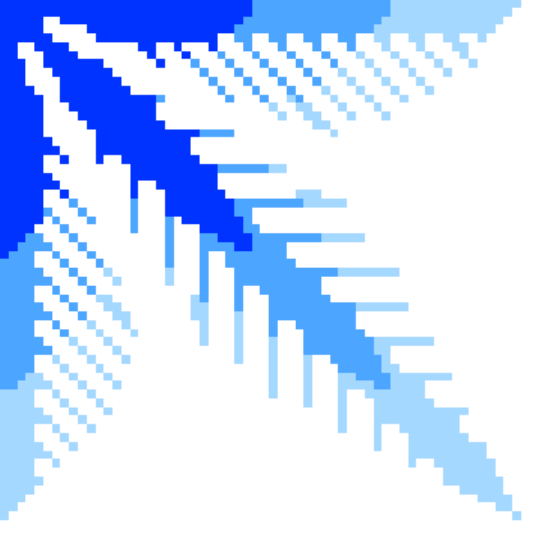}
        \subcaption{Crystal growth simulation by CGNE.}
        \label{fig:snowflake-rollout-ours}
    \end{minipage}
    \caption{Illustration of the crystallization simulation process. To account for the crystal's symmetries, we first convert the hexagonal domain on which the snowflake lives to a square grid representation. Then, we simulate a quadrant of the square grid, corresponding to one-third of the hexagonal grid. The crystal growth process by CGNE produces a crystal which is qualitatively very similar to the one produced by the numerical LCA method.
    }
    \label{fig:snowflake-rollout-frontpage}
\end{figure*}

Solidification is the process of a substance transitioning from a liquid or gaseous state to a solid state, which is evident in various systems, including the crystallization of metals and the growth of snow crystals from water vapor, illustrated in Figure~\ref{fig:snowflake-rollout-frontpage}. The interplay of highly sensitive nonequilibrium and nonlinear processes as well as stochastic effects result in complex structures emerging during the solidification process. As such, studying these processes not only has significant practical implications in fields like materials science and metallurgy \cite{2022metal_snowflakes}, but can also help us to understand the mechanisms behind emergent complexity in general.

In particular, this complexity manifests itself at the mesoscopic scale: while at the atomic scale one might see atoms arrange themselves into a well-organised lattice, and at the macroscopic scale alloys solidify to an ingot, intricate pattern formations caused by the polycrystalline solidification process arise at the mesoscopic scale. 

Numerical simulation is a powerful tool to enhance our understanding of these processes. Various numerical methods have been developed for crystal formation, with many able to model stochastic growth to capture the diversity observed in real-world complex solidification processes. Computational models based on front-tracking~\cite{yokoyama1990pattern, libbrecht1999cylindrically} and phase-field~\cite{karma1998quantitative, boettinger2002phase, granasy2004modelling, biswas2022solidification} techniques contributed valuable insights for multigrain and dendritic growth simulations, but occasionally struggled with dynamic and numerical instabilities. In particular, such instabilities arise when modeling combined faceted and dendritic growth, for example as observed in snow crystals~\cite{Barrett_2012, 2008libbrecht_phys_ca}. More recently, local cellular automaton (LCA) based models~\cite{gravner2006packard, raghavan2007modeling, gravner2008modeling, 2008libbrecht_phys_ca, gravner20093d, kelly2013physical, svyetlinchnyy2013modeling} have advanced the field, capturing previously unseen morphological features for combined faceted and dendritic growth~\cite{Libbrecht_Snow_Crystals}. However, these numerical models are computationally expensive, due to the high spatiotemporal resolution required to uphold simulation accuracy in the sensitive crystallization process.

Recently, scaling numerical simulations with neural networks has seen a surge in interest from the scientific machine learning community, with successful applications in climate science~\cite{nguyen2023climax, gao2023prediff, bonev2023sfno}, nuclear fusion~\cite{poels2023fast, Gopakumar2024PlasmaSurrogate}, and fluid dynamics~\cite{gupta2022pdearena, li2023geometryinformed}. Similarly, various approaches have been proposed in the context of crystal growth modeling. One significant branch of research focuses on Molecular Dynamics (MD) simulation of crystallization~\cite{zhang2024unveiling, jarry2023medium}. While such MD methods can impressively scale to simulations of up to 1 million atoms, they are still operating at the microscopic scale, and a simulation at the mesoscopic scale would require many orders of magnitude more atoms to be simulated. Alternatively, \citet{ren2022numerical}, \citet{nomoto2019solidification} and \citet{xue2022physics} accelerate phase field simulations directly at the mesoscopic scale. However, in this work we consider a stochastic crystallization process. As these methods are deterministic, they do not apply to our setting. More broadly, probabilistic neural simulation methods have been proposed, but these do not straightforwardly apply to the discrete crystal growth problem~\cite{cachay2023dyffusion, yang2023denoising, bergamin2024guided}, or fail to reliably capture the sensitive and stochastic relationship between environmental parameters and crystal morphology~\cite{minartz2023equivariant}.

Although these prior works provide interesting and relevant methods for crystal growth or probabilistic neural simulation in general, no method exists for efficient probabilistic crystal growth simulation at the mesoscopic scale. In this work, we introduce a neural simulator model that greatly accelerates the simulation of mesoscopic scale crystallization processes. Our main contributions are as follows:

\begin{itemize}
    \item We propose the Crystal Growth Neural Simulator (CGNE),\footnote{ \url{https://github.com/poltimmer/CGNE}} a probabilistic model for simulating crystal growth processes at the mesoscopic scale.
    \item We observe that existing training techniques fail to enable the model to reliably capture the joint distribution over environmental parameters and crystal morphologies. We diagnose Latent Variable Neglect (LVN) as the root cause of the problem and introduce a technique, called samplewise decoder dropout, that systematically prevents LVN. 
    \item We construct a new snow crystal growth dataset that captures the intricate relationships between environmental parameters and crystal morphology.
    \item We show that CGNE outperforms a state-of-the-art method for probabilistic neural simulation~\cite{minartz2023equivariant}. Specifically, we demonstrate that introducing decoder dropout improves simulation diversity and sample quality by preventing LVN, which has significant implications for neural simulation problems beyond crystal growth.
\end{itemize}

\section{Approach}
\subsection{Problem Formulation}
At time $t$ the system is in state $(x_t, y)$, where $x_t$ is a binary function on a fixed grid and $y$ is a real-valued vector of static environmental parameters. The system evolution is specified with a sequence of states $(x_{0:T}, y)$. Our goal is to fit a simulation model $p_\theta$ to the data by maximizing the log-likelihood over the training set $\left\{\left(x_{0:T}, y \right)_i \right\}_{i=1}^N$:
\begin{equation}
    \frac{1}{N} \sum_{i=1}^N \log p \left(\left(x_{0:T}, y\right)_i\right)\text{.}
\end{equation}
As LCA-based numerical simulators are Markovian, and we assume that the distribution over initial conditions and environmental parameters is known, this is equivalent to maximizing \autoref{eq:autoregressive} with respect to $\theta$:
\begin{equation}
    \label{eq:autoregressive}
    \frac{1}{N} \sum_{i=1}^N \sum_{t=0}^{T-1} \log p\left(\left(x_{t+1}\mid x_t, y\right)_i\right)\text{.}
\end{equation}
After training on~\autoref{eq:autoregressive}, we can generate new samples by applying $p\left(\left(x_{t+1}\mid x_t, y\right)_i\right)$ autoregressively.

\begin{figure*}[ht]
    \centering
    \begin{subfigure}[t]{0.49\linewidth}
        \includegraphics[width=1\linewidth]{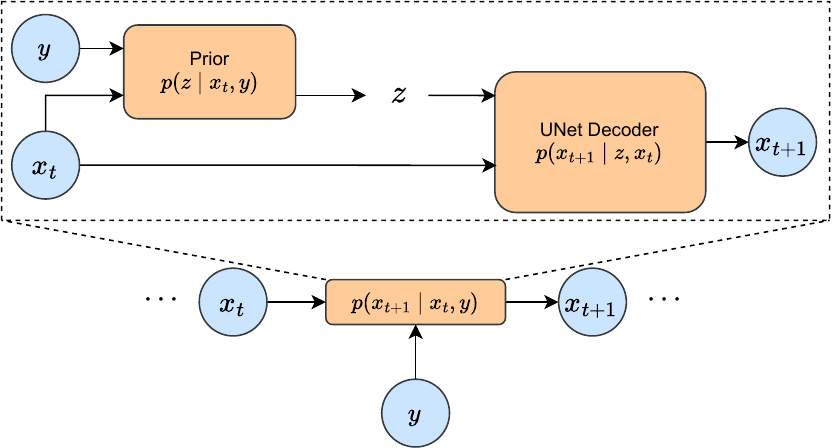}
        \caption{Sample pass of the model.}
        \label{fig:model-sample}
    \end{subfigure}
    \hfill
    \begin{subfigure}[t]{0.49\linewidth}
        \centering
        \includegraphics[width=1\linewidth]{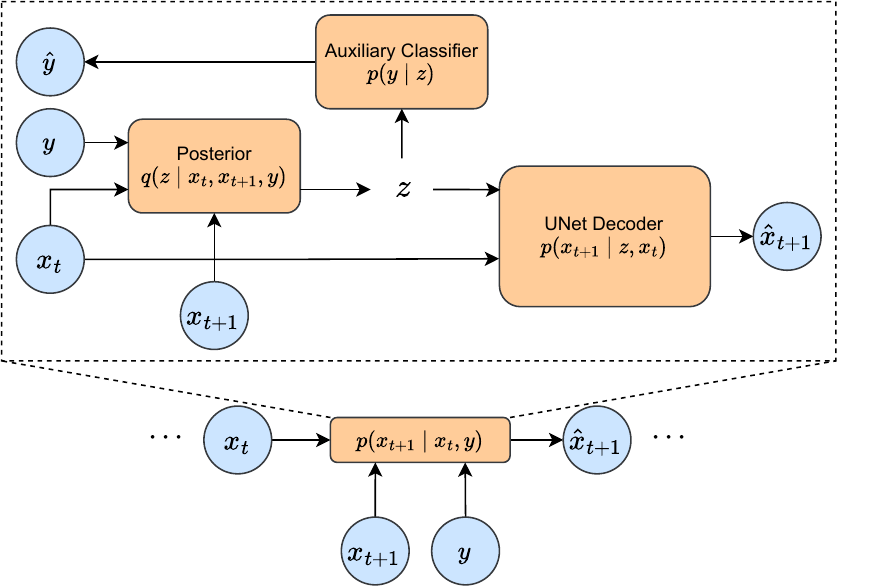}
        \caption{Training pass of the model.}
    \label{fig:model-train}
    \end{subfigure}
    \caption{CGNE model overview.}\label{fig:model-overview}
\end{figure*}

\subsection{Model Design}
We aim to develop a model capable of producing realistic simulation trajectories with good sample efficiency. Toward this goal, we build our model using the PNS framework~\cite{minartz2023equivariant} for autoregressive probabilistic neural simulation. It follows a Conditional Variational Autoencoder (CVAE) structure, where the conditional prior is conditioned on the simulation state $x_t$, and the posterior is conditioned on both the current and the next state $x_t, x_{t+1}$. However, PNS does not support conditioning on environmental parameters. To address this, we choose to additionally condition both the prior and posterior on the environmental parameters $y$, which gives a conditional prior $p(z\mid x_t, y)$, and a posterior $q(z\mid x_t, x_{t+1}, y)$. An overview is given in \autoref{fig:model-overview}.

\begin{figure*}[ht]
    \centering
    \includegraphics[width=0.9\linewidth]{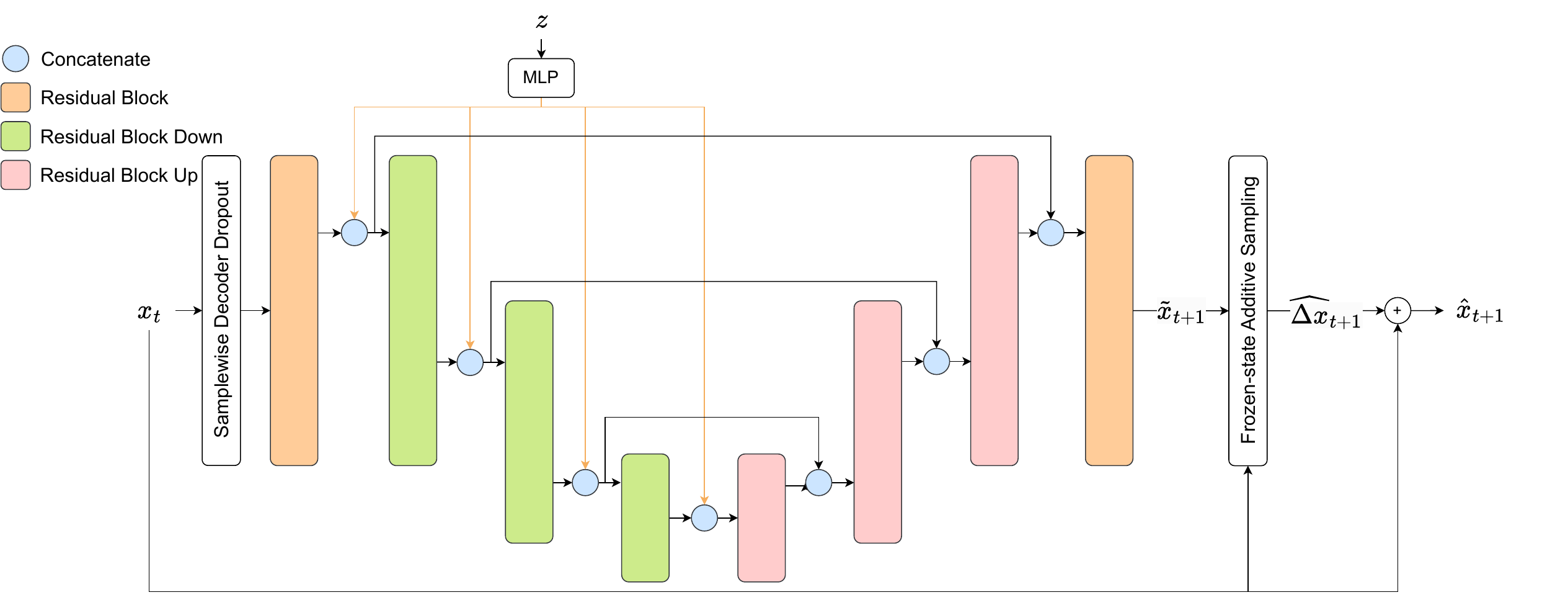}
    \caption{Overview of CGNE's decoder, including the samplewise decoder dropout and frozen-state additive sampling strategies.}
    \label{fig:decoder-architecture}
\end{figure*}

CGNE is fully convolutional and thus preserves the local translation symmetries of the data domain. At the implementation level, the latent space $z$ is also a grid instead of a vector, and the decoder is a fully convolutional UNet, as detailed in \autoref{fig:decoder-architecture}. The latent variable is passed through an MLP and then concatenated with the input of every residual block of the UNet.

We only consider the solidification processes, so no melting or sublimation is modeled. We incorporate this property into the model through frozen-state additive sampling, as illustrated in \autoref{fig:decoder-architecture}, where the solidified portion from the previous step is preserved, and the model’s predictions are added only to the non-solidified pixels, ensuring a physically accurate solidification process.

Finally, we use a small auxiliary classifier $p(y\mid z)$ that reconstructs $y$ from the latent variable during training, improving the representation of the environmental parameters in the latent space~\cite{ilse2020diva}.

\subsection{Training Strategies}
\paragraph{Latent variable neglect.}

Because the decoder is conditioned on the previous simulation state $x_t$, and differences between consecutive states ($x_t$ and $x_{t+1}$) are very small inferring a reasonable estimate for $x_{t+1}$ is possible to a large degree without the use of $z$. Consequently, the gradients from $z$, which have a relatively low signal-to-noise ratio, are easily overshadowed by the gradients from $x_t$, with their relatively high signal-to-noise ratio. This effect is amplified by the skip connections present in the UNet architecture used by the decoder. This phenomenon causes the model to neglect the latent variables. Notably, although the one-step-ahead estimates of such a deterministic model might be reasonable, the lack of accumulation of randomness results in a severe underestimation of sample diversity at the full trajectory rollout level. Even though, LVN is similar to posterior collapse, where the posterior $q(z \mid x_t, x_{t+1}, y)$ collapses to the prior $p(z \mid x_t, y)$, in this case, the environmental parameters $y$ are also ignored. In other words, the conditional prior also collapses to an uninformative prior.

To remedy this issue, we use a samplewise dropout on the direct connection from $x_t$ to the decoder. This approach is similar to word dropout used by Bowman et al.~\cite{bowman2016generating}. The dropout probability is set on a schedule, starting at 1 for the first short part of the training, then dropping linearly to 0.1 over a few epochs, after which it remains at 0.1 for the remainder of the training. During inference, the dropout is disabled. This training strategy reinforces the flow of information through the latent variables, which solidifies the stochastic pathway and consequently significantly improves the model's ability to model realistic and diverse trajectories.

Furthermore, we consider beta annealing~\cite{bowman2016generating} and the free bits trick~\cite{kingma2016improved}. These techniques prevent posterior collapse, a failure mode where the KL term of the VAE loss dominates the reconstruction term, preventing the model encoder from learning useful representations. Although these techniques have been shown to improve both diversity and simulation quality in PNS~\cite{minartz2023equivariant}, the improvement is relatively minor compared to the simple yet effective decoder dropout technique.

\paragraph{Collapse to the identity function.}
Another issue arises when modeling a system characterized by gradual differences in subsequent simulation states with a Markov model. When differences between subsequent states are too small, the model can struggle to learn the actual dynamics of the simulation and instead tends to resort to the strong local minimum of an identity function. In this failure mode, model simulations tend to deadlock at some point in the trajectory. Therefore, to improve simulation efficiency and prevent this failure mode, we downsample the temporal resolution, increasing the observed differences between subsequent simulation states.

\section{Experiments}
\subsection{Experiment Setup}
\paragraph{Dataset.}
We choose to validate the model on snow crystal growth, as it exemplifies a complex crystallization process, which stems from the underlying combination of faceted and dendritic growth. We generate a dataset by implementing Janko Gravner and David Griffeath's stochastic LCA algorithm for snow crystal growth \cite{gravner2008modeling}. This LCA model simulates the growth of a snowflake from a single seed crystal and has seven tuneable environmental parameters that influence the growth and final morphology of the crystal. We fix six of these parameters and vary the parameter $\rho$, which corresponds to the ambient vapor saturation. Specifically, we sample $\rho$ from a uniform distribution within the reference range of that parameter of 0.35 - 0.65. This parameter is most influential on the shape of the snow crystal and therefore leads to a wide range of morphologies when varied.

\paragraph{Metrics and baselines.}

Because the growth process is stochastic, snow crystals generated from the same environmental parameters can differ significantly at the pixel level. While low $\rho$ tends to generate thin crystals with few side branches, and a high $\rho$ typically generates a thick crystal with many side branches, the exact location where these branches attach to the crystal can vary within the same $\rho$. Consequentially, a simple pixel-wise accuracy metric is not sufficiently useful for model validation. Instead, we opt to validate the model by investigating the high-level morphological properties of the generated structure.

Specifically, we investigate if the joint distribution $p(a, b \mid \rho)$ of the model's generated snowflakes, conditioned on the environmental parameter $\rho$, matches the joint conditional distribution of the numerical LCA simulator $p_\text{LCA}(a, b \mid \rho)$. Where $a$ is the area of the snowflake and $b$ is the boundary length. We quantify the match of these distributions using the expected value of the Wasserstein distance between the conditional distributions, where the expectation is taken over the environmental parameter $\rho$:
\begin{equation}\label{eq:wasserstein-aggregated}
    \text{EWD} = \mathbb{E}_{\rho \sim p(\rho)} \left[W_2\left(p_\theta(a, b \mid \rho), p_\text{LCA}(a,b \mid \rho) \right)\right]\text{.}
\end{equation}
In~\autoref{eq:wasserstein-aggregated}, $W_2$ denotes the type-2 Wasserstein distance. Intuitively, the EWD is small if the conditional distributions $p(a,b \mid \rho)$ overlap well with the ground-truth distribution $p_\text{LCA}(a,b \mid \rho)$.
As a second metric, we consider the Evidence Lower Bound (ELBO) of the model on the held-out data. The ELBO is a lower bound on the log-likelihood, defined in~\autoref{eq:autoregressive}, where a higher value indicates that the model allocates more probability mass to real samples, and suggests a better fit to the data. As a baseline, we take PNS~\cite{minartz2023equivariant}, a recent state-of-the-art model for probabilistic simulation of dynamical systems.

\subsection{Results}
\paragraph{Qualitative results.}

\begin{figure}[t]
    \centering
    \includegraphics[width=1\linewidth]{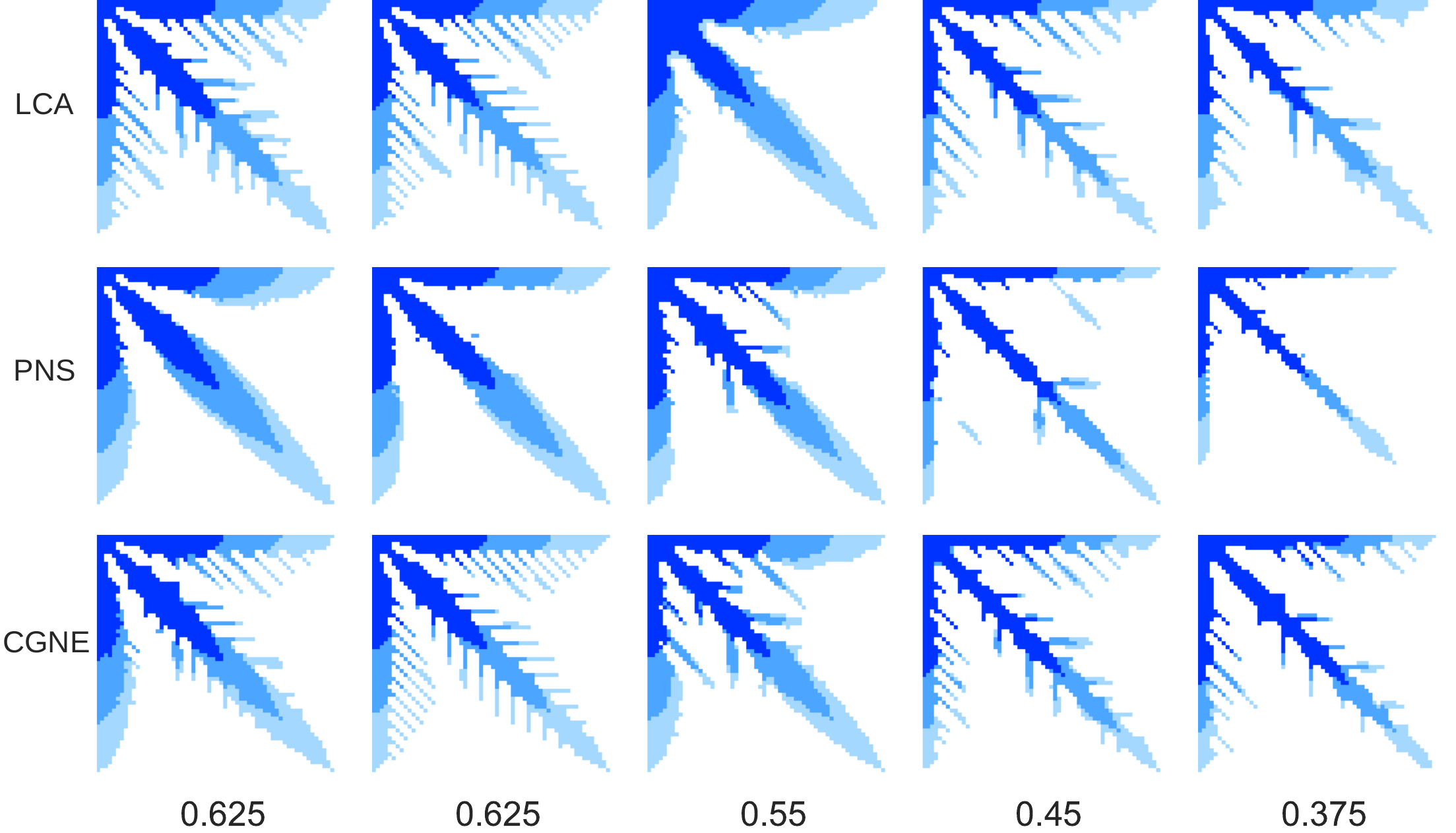}
    \caption{Qualitative comparison of simulated trajectories for different values of $\rho$. Three stages of growth are displayed with different shades of blue.}
    \label{fig:rollouts}
\end{figure}
\begin{figure*}[h]
    \centering
    \begin{subfigure}[b]{0.24\linewidth}
        \centering
        \includegraphics[width=\linewidth]{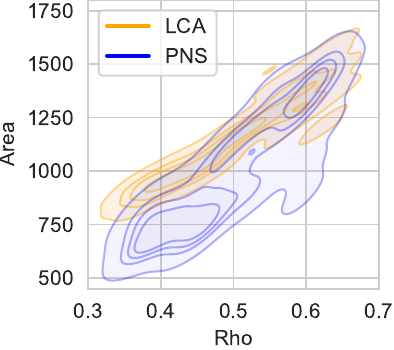}
        \caption{Area, PNS and LCA.\newline}
        \label{fig:contour-area-pns}
    \end{subfigure}
    \hfill
    \begin{subfigure}[b]{0.24\linewidth}
        \centering
        \includegraphics[width=\linewidth]{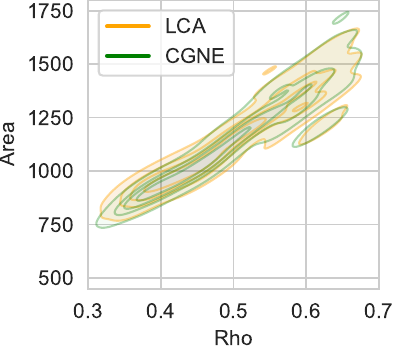}
        \caption{Area, CGNE and LCA.\newline}
        \label{fig:contour-area-ours}
    \end{subfigure}
    \hfill
    \begin{subfigure}[b]{0.24\linewidth}
        \centering
        \includegraphics[width=\linewidth]{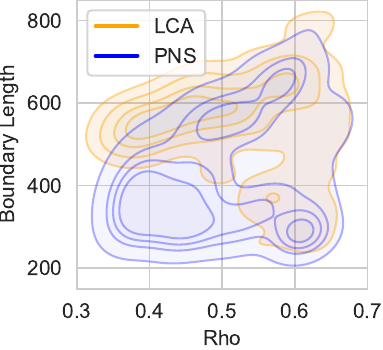}
        \subcaption{Boundary length, PNS and LCA.}
        \label{fig:contour-bl-pns}
    \end{subfigure}
    \hfill
    \begin{subfigure}[b]{0.24\linewidth}
        \centering
        \includegraphics[width=\linewidth]{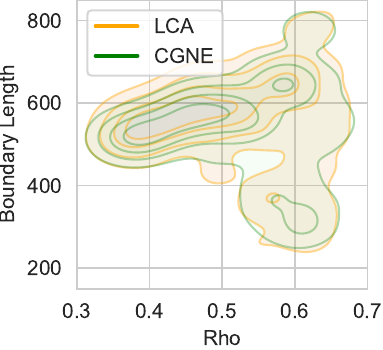}
        \subcaption{Boundary length, CGNE and LCA.}
        \label{fig:contour-bl-ours}
    \end{subfigure}
    \caption{Joint distributions over morphological crystal properties and environmental parameters for PNS and CGNE, compared with the actual joint distribution produced by LCA.}
    \label{fig:morphology-distribution-comparison}
\end{figure*}

\autoref{fig:rollouts} illustrates a set of simulated snowflakes comparing our CGNE method to the LCA ground truth and simulations using the PNS model. For five values of $\rho$, we simulate a full snowflake using the three models. In the snowflakes produced by the LCA simulator, we observe a trend of thinner, less voluminous snowflakes as $\rho$ decreases. Intuitively, this can be explained by $\rho$ corresponding to lower atmospheric water vapor saturation. Furthermore, we observe that the LCA simulator can produce two different-looking snow crystals for the same $\rho$. As the ground truth data-generating process is stochastic, achieving this level of diversity is an important aspect of a good neural simulation model for this setting. PNS struggles to capture this diversity, as displayed by the two snowflakes generated from $\rho = 0.625$ being very similar. Without samplewise decoder dropout, the model suffers from latent variable neglect and edges closer to a deterministic model. 

The sample quality of the simulated snowflakes by PNS is also noticeably worse than for CGNE. PNS tends to struggle with the formation of thin branches and finer details. This often results in PNS-simulated snowflakes lacking thin side branches, especially for lower values of $\rho$, and it results in distortions and artifacts in the form of floating crystal pieces, which are physically impossible in the ground truth data-generating process. Conversely, snowflakes generated by CGNE look complete and similar to the ground truth for all values of $\rho$. The model shows no artifacts and has no issues with thinner branches, even for low values of $\rho$. The snowflakes visualized in the plot are not identical between the LCA and CGNE models, but this is expected as the models are stochastic. Importantly, the snowflakes generally share a similar style and do not show any distortions.

\paragraph{Distribution overlap.}

\begin{table}[t]
    \centering
    \caption{Quantitative results. The lower expected Wasserstein distance EWD indicates that simulations generated by CGNE match the ground-truth data-generating process better than PNS, while the higher ELBO on the test set indicates a more strongly peaked probability distribution on real crystallization processes.}
    \label{tab:results}
    \begin{tabular}{lccc}
        \toprule
        Method & EWD $\downarrow$ & ELBO $\uparrow$ \\
        \midrule
        PNS & 202.8 & -0.0670 \\
        CGNE & \textbf{43.8} & \textbf{-0.0428} \\
        \bottomrule
    \end{tabular}
\end{table}
We now investigate to what extent the distribution over snowflake morphologies generated by CGNE and PNS aligns with the numerical simulator. Figure~\ref{fig:morphology-distribution-comparison} shows the joint distribution over the environmental parameter $\rho$ and the crystal area and boundary length, comparing PNS and CGNE with LCA. We observe that the distributions generated by CGNE overlap very well with the ground truth, capturing all the relevant modes as well as the overall correlation, as seen in Figures~\ref{fig:contour-area-ours} and~\ref{fig:contour-bl-ours}. In contrast, the distributions generated by PNS do not show the correct overall correlation, shown in Figure~\ref{fig:contour-area-pns}, and do not capture the modes well, as demonstrated in Figure~\ref{fig:contour-bl-pns}. Specifically, snow crystals generated by PNS have lower boundary length and lower area, especially at lower $\rho$. This can be attributed to PNS struggling with fine branch formation, which occurs primarily at these low $\rho$ values.

Quantitatively, this difference in correlation is also observed, as shown in \autoref{tab:results}. The expected Wasserstein distance is significantly better for CGNE compared to PNS. This indicates that the distribution over crystal morphologies generated by CGNE overlap better with those generated by LCA. Furthermore, the ELBO of CGNE is also better than that of PNS, reflecting a higher lower bound on the log-likelihood over the test data as expressed by Equation~\ref{eq:autoregressive}. This means that CGNE allocates more probability mass on real data, suggesting a better fit.

\begin{figure}[t]
    \centering
    \includegraphics[width=1\linewidth]{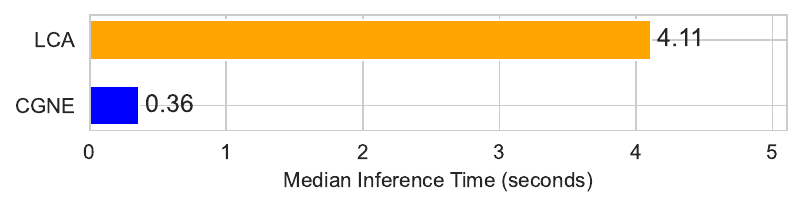}
    \caption{Median inference time of a full simulation trajectory for LCA and CGNE.}
    \label{fig:speed-comparison}
\end{figure}

\paragraph{Inference speedup.}
CGNE significantly increases inference efficiency over our GPU-accelerated implementation of the LCA model. On average, the inference of a snow crystal growth trajectory is 11x faster with CGNE compared to the LCA model as shown in \autoref{fig:speed-comparison}. As the speed of inference is inversely proportional to the temporal resolution, this speed increase is flexible and can be even more significant depending on the use case.

\section{Conclusions and Future Work}
We propose a model architecture CGNE, that greatly accelerates probabilistic simulations of crystallization processes at the mesoscopic scale while maintaining excellent simulation accuracy. We identify Latent Variable Neglect as a significant challenge for such autoregressive models, which we resolve with samplewise decoder dropout. As a result, CGNE significantly improves the simulation accuracy and diversity compared to PNS, a recent state-of-the-art model for probabilistic simulation of dynamical systems. 

An interesting avenue for future research would be implementing a mechanism that physically prevents distortions in the form of floating crystal pieces. While these distortions are not an issue in the trained CGNE model, enforcing this property of crystal growth into the model directly could improve training time, or could enable the model to uphold accurate simulations over greater spatial resolutions. Finally, it would be interesting to apply CGNE to more crystallization processes, for example in metallurgy, to investigate how the method generalizes to different types of crystallization.


\bibliography{main}
\bibliographystyle{icml2024}

%%%%%%%%%%%%%%%%%%%%%%%%%%%%%%%%%%%%%%%%%%%%%%%%%%%%%%%%%%%%%%%%%%%%%%%%%%%%%%%
%%%%%%%%%%%%%%%%%%%%%%%%%%%%%%%%%%%%%%%%%%%%%%%%%%%%%%%%%%%%%%%%%%%%%%%%%%%%%%%
% APPENDIX
%%%%%%%%%%%%%%%%%%%%%%%%%%%%%%%%%%%%%%%%%%%%%%%%%%%%%%%%%%%%%%%%%%%%%%%%%%%%%%%
%%%%%%%%%%%%%%%%%%%%%%%%%%%%%%%%%%%%%%%%%%%%%%%%%%%%%%%%%%%%%%%%%%%%%%%%%%%%%%%
% \newpage
% \appendix
% \onecolumn

% You can have as much text here as you want. The main body must be at most $8$ pages long.
% For the final version, one more page can be added.
% If you want, you can use an appendix like this one.  

% The $\mathtt{\backslash onecolumn}$ command above can be kept in place if you prefer a one-column appendix, or can be removed if you prefer a two-column appendix.  Apart from this possible change, the style (font size, spacing, margins, page numbering, etc.) should be kept the same as the main body.
%%%%%%%%%%%%%%%%%%%%%%%%%%%%%%%%%%%%%%%%%%%%%%%%%%%%%%%%%%%%%%%%%%%%%%%%%%%%%%%
%%%%%%%%%%%%%%%%%%%%%%%%%%%%%%%%%%%%%%%%%%%%%%%%%%%%%%%%%%%%%%%%%%%%%%%%%%%%%%%

\end{document}